\documentclass[letterpaper, 10 pt, conference]{ieeeconf}  

\IEEEoverridecommandlockouts                              

\usepackage{graphicx}
\usepackage{amsmath}
\usepackage{float}
\usepackage{soul}
\usepackage{xcolor} 
\usepackage{placeins}

\sethlcolor{cyan}  

\title{\LARGE \bf
Policy Learning for Social Robot-Led Physiotherapy
}

\author{Carl Bettosi$^{1, 2}$, Lynne Baillie$^{1}$, Susan Shenkin$^{3}$ and Marta Romeo$^{1}$
\thanks{This work was funded by the UK Engineering and Physical Sciences Research Council (Grant Number: EP/S023208/1)}
\thanks{$^{1}$Dept. of Mathematical and Computer Sciences, Heriot-Watt University, UK {\tt\small cb54@hw.ac.uk, l.baillie@hw.ac.uk, m.romeo@hw.ac.uk}}%
\thanks{$^{2}$School of Informatics, University of Edinburgh, UK}%
\thanks{$^{3}$Ageing and Health, and Advanced Care Research Centre, Usher Institute,
University of Edinburgh, UK {\tt\small susan.shenkin@ed.ac.uk}}%
}
\begin{document}

\maketitle
\thispagestyle{empty}
\pagestyle{empty}

\begin{abstract}
Social robots offer a promising solution for autonomously guiding patients through physiotherapy exercise sessions, but effective deployment requires advanced decision-making to adapt to patient needs. A key challenge is the scarcity of patient behavior data for developing robust policies. To address this, we engaged 33 expert healthcare practitioners as patient proxies, using their interactions with our robot to inform a patient behavior model capable of generating exercise performance metrics and subjective scores on perceived exertion. We trained a reinforcement learning-based policy in simulation, demonstrating that it can adapt exercise instructions to individual exertion tolerances and fluctuating performance, while also being applicable to patients at different recovery stages with varying exercise plans.
\end{abstract}

\section{INTRODUCTION}

Adherence to physical exercise programs is crucial for rehabilitating individuals who have sustained injuries such as fractures or strokes, significantly improving physical outcomes while delivering broader social and economic benefits~\cite{patel2020estimated}. However, healthcare workforce shortages in many countries mean patients can go weeks or months without in-person physiotherapy assessments, meaning they often rely on generic paper or video-based exercise guides~\cite{suero2025effectiveness}. While recent technological advances in the physiotherapy space show promise, many solutions remain reliant on constant human oversight, creating an increase in staff workload~\cite{biggs2025physiotherapy} and training~\cite{martinsen2024usage}, highlighting the need for autonomous, accessible approaches that can provide personalized rehabilitation support in community settings.


Social robots have demonstrated the potential to autonomously deliver exercise sessions~\cite{winkle2020situ, pulido2019socially}, adapting instructions based on real-time user feedback while leveraging their physical presence to enhance engagement beyond screen-based agents~\cite{vasco2019train}. However, developing autonomous control policies is challenging, as patients differ in physical ability, with performance fluctuating daily, requiring the system to recognize and adapt to both good and bad days.

In our previous work, we co-designed a standardized upper-limb exercise session with physiotherapists~\cite{bettosidesigning}. In this work, we leverage our acquired knowledge and develop a social robot prototype capable of delivering the physiotherapy session one-on-one with patients through verbal, visual, and physically demonstrated instruction. To enable autonomous decision-making during these sessions, we train a reinforcement learning (RL) agent in a simulated environment that incorporates a dynamic patient behavior model. 

In human-robot interaction (HRI), online learning can be costly due to the significant time investment required for interactions with humans~\cite{akalin2021reinforcement}. Additionally, modeling end-user behavior in software simulations is inherently challenging, as creating large-scale, representative datasets for training these models can be both difficult and time-consuming~\cite{yu2018towards}. To address these challenges, we collaborate with expert clinicians to integrate insights that capture a broad spectrum of patient behaviors. Our goal is to generate a robot control policy that can adaptively instruct the most optimal number of exercise repetitions in real-time based on:

\begin{itemize}
    \item Patient-specific exertion thresholds, allowing for personalized intensity adjustments during sessions
    \item Session-to-session variations in patient performance
    \item Individual exercise plans tailored to each patient's physical capabilities and rehabilitation goals
\end{itemize}

\section{BACKGROUND AND RELATED WORK}

There is a notable body of work exploring social robots in the physical exercise domain. These robots often serve a variety of functions: as motivational companions~\cite{swift2015effects}, as a means of offering corrective feedback on exercises~\cite{bogliolo2020robot}, or as instructors actively leading exercise sessions~\cite{ross2024implementation}. Machine learning-based approaches allowing for personalization in response to session dynamics have been a key focus in these applications. For example, Tapus et al. uses RL to optimize proxemics, speed, and vocal content to introverted/extroverted personality types~\cite{tapus2008user}. Winkle et al. demonstrated how an interactive machine learning approach, trained with real-world data, improved autonomous robot behavior for exercise sessions, though the model struggled with timing~\cite{winkle2020situ}. Pulido et al. developed a social robot for pediatric upper-limb rehabilitation, using automated planning to adapt exercises based on user performance, but found a high reliance on expert intervention~\cite{pulido2019socially}.





Although machine learning is helping to achieve personalization in HRI, learning to adapt to the nuances of human task behavior—whether through direct training or simulation—remains highly challenging. This is particularly true in sensitive domains like healthcare, where acquiring patient behavior data is difficult. Researchers often rely on passive task observations~\cite{ross2021observing} or interactive machine learning with experts~\cite{winkle2020situ}, but these methods can be resource-intensive. An alternative approach is to use a sample of representative interactions to inform user behavior models used in simulation~\cite{andriella2019learning}. For instance, Tsiakas et al.~\cite{tsiakas2018task} used RL in a social robot-led cognitive learning task, adapting difficulty based on user performance and engagement, where simulated users were derived from clustered user types. Similarly, Stolarz et al. applied RL to an robot-led assistive therapy task for autism spectrum disorder, using a learned user model fitted to a dataset of sampled human behavior~\cite{stolarz2024learning}. In these examples, however, non-end users were used to inform user models, inevitably leading to a gap between simulated and real-world performance. Where collecting end-user behavior is infeasible, a more rigorous approach may involve leveraging task experts with experience in end-user behavior.

\section{A SOCIAL ROBOT EXERCISE INSTRUCTOR}

\begin{figure}[t]
    \centering
    \includegraphics[width=\linewidth]{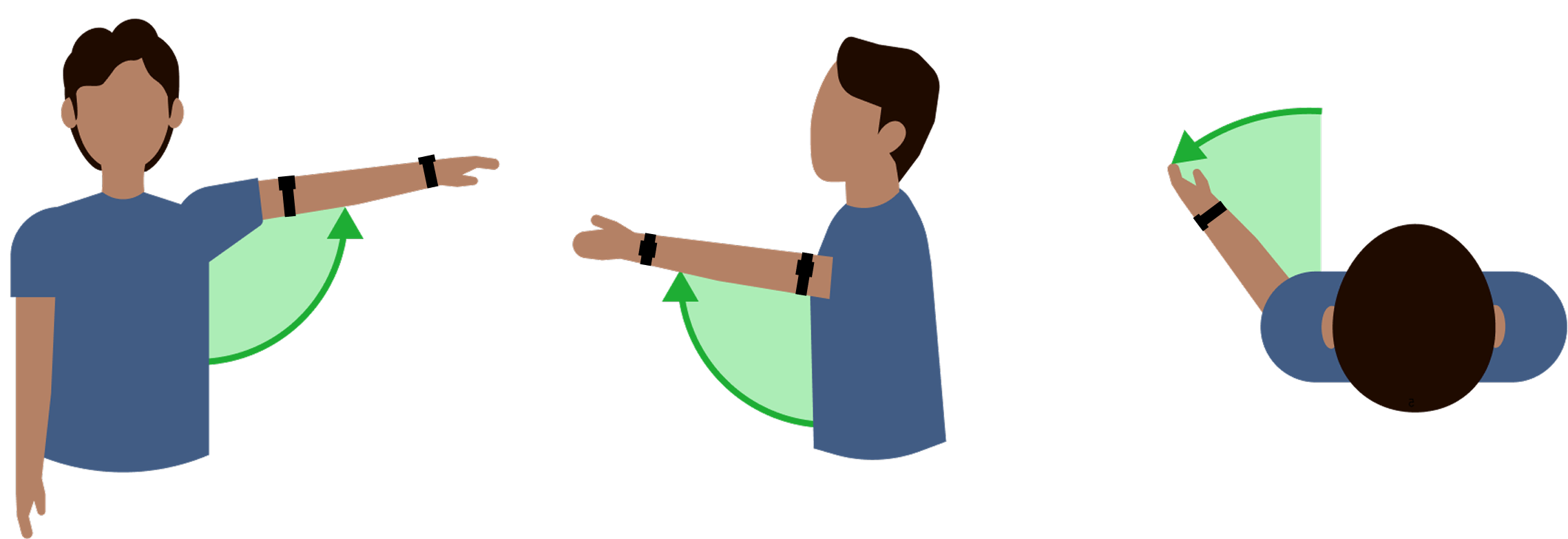}
    \caption{The three upper limb exercises incorporated into our prototype. Left: arm abduction (front view), center: arm flexion (side view), right: arm external rotation (top view).}
    \label{fig:exercises}
\end{figure}

\begin{figure}[t]
    \centering
    \includegraphics[width=\linewidth]{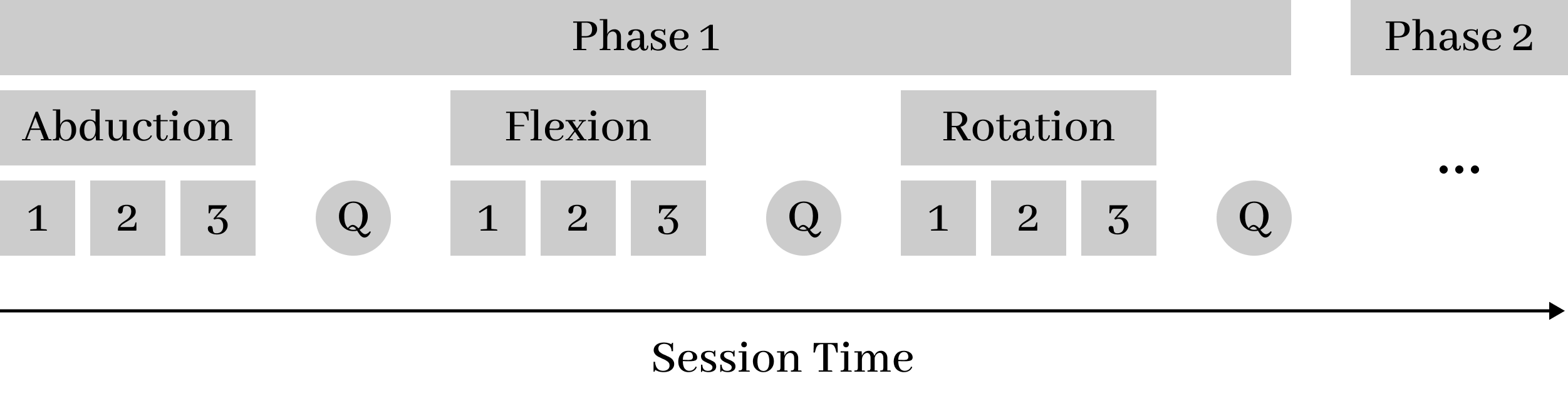}
    \caption{The structure of our robot's guided exercise sessions as informed through our expert-led design~\cite{bettosidesigning}.}
    \label{fig:session-structure}
\end{figure}

    \subsection{Robot Platform}
    
    We developed a prototype on the ARI platform by PAL Robotics~\cite{cooper2020ari}, a 165cm tall humanoid robot with rich multimodal interaction capabilities that has shown strong suitability in real-world healthcare applications~\cite{cooper2022towards}.
    The robot's task is to guide a patient through an exercise session that can last up to 20 minutes. The robot communicates feedback and exercise instructions (e.g., `Do 6 reps of arm rotations') using verbal utterances and screen captions. The instructions are supplemented with still animations displayed on the screen (similar to those in Figure \ref{fig:exercises}) and demonstrations performed through the robot's upper limbs. Each limb has 2 degrees of freedom (DoF) in the shoulder, and 1 DoF in the upper arm, elbow, and wrist, allowing for accurate representations of the exercises. The design of these interactions was shaped by prior design workshops and interviews with physiotherapists and occupational therapists~\cite{bettosi2024systematic, bettosidesigning}. 
    
    \subsection{Standardized Upper Limb Exercise Session}

    The three exercises instructable the prototype (abduction, flexion, rotation), illustrated in Figure \ref{fig:exercises}, are popular in standard UK National Health Service shoulder rehabilitation materials. They provide good coverage of shoulder movement, are adaptable for sitting or standing, and easily sensed by the robot, supporting our goal of autonomy. To accommodate long-term rehabilitation progression, each exercise can be performed using one of three modes, allowing for nine exercise variations:
    \begin{itemize}
        \item Assistive: Using props or the unaffected limb to support movement
        \item Active: Performing the movement using only the affected limb to improve range of motion (ROM)
        \item Resistive: Incorporating resistance bands to build strength
    \end{itemize}
    
    Our robot follows the same structure for each session, shown in Figure \ref{fig:session-structure}. This consists of two overarching phases depending on patient ability (\{assistive, active\} or \{active, resistive\}). \{active, resistive\} is excluded, as patients ready for strengthening typically no longer require assistance. In each phase, patients complete three exercise blocks—abduction, flexion, and rotation—with three sets per block. The robot controls the number of repetitions in each set. After each block, the robot asks the patient for feedback on their perceived exertion. This standardized session design creates a consistent framework that simplifies the robot's learning process but maintains alignment with clinical advice~\cite{bettosidesigning}.

    \begin{figure}[t]
        \centering
        \includegraphics[width=6cm]{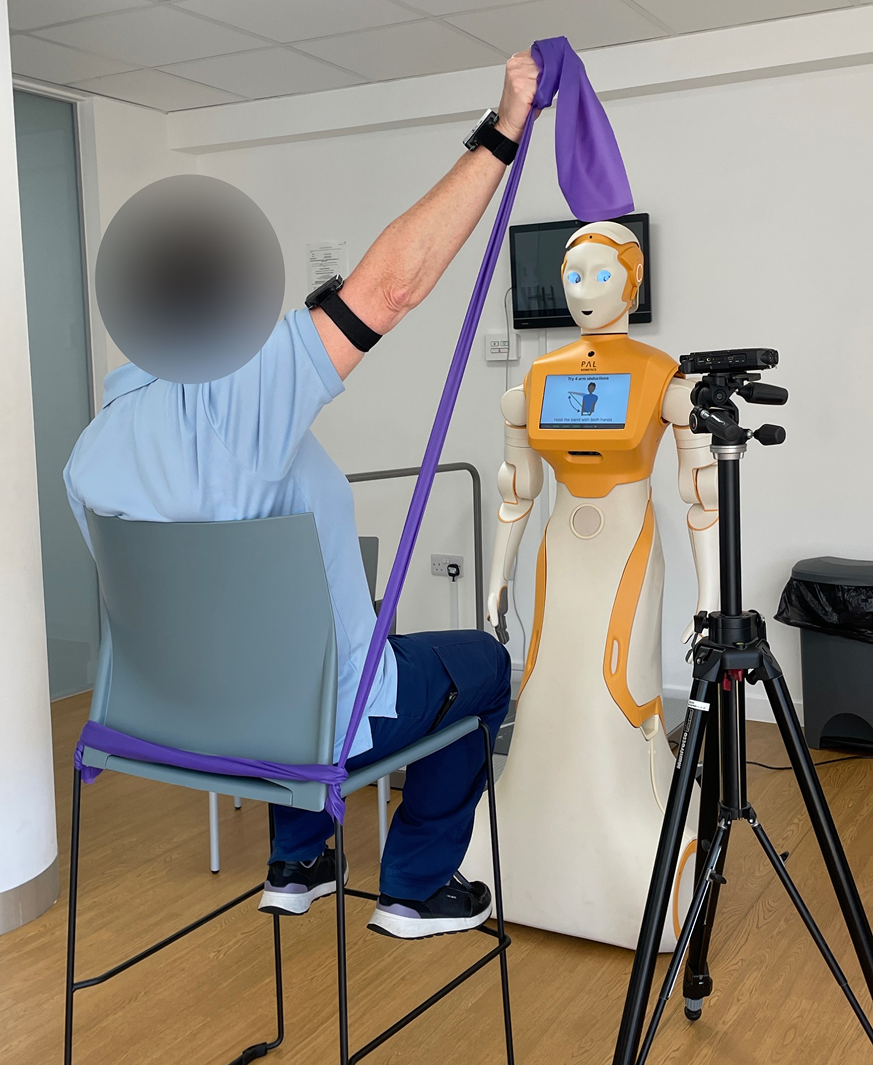}
        \caption{A participant engaging with our robot during a data collection session. The participant is performing a resistive abduction exercise as they roleplay a high ability level.}
        \label{fig:physio-robot}
    \end{figure}
    
    \subsection{Exercise Perception} \label{subsec:exercise_perception}

    \subsubsection{Repetition count}
    To assess patients' exercise performance, two wearable Inertial Measurement Units (IMUs)\footnote{https://x-io.co.uk/x-imu3/} are fitted to the affected arm, as shown in Figure \ref{fig:physio-robot}. IMUs have been proven effective for capturing upper limb exercise data~\cite{lee2024deep}, avoiding the limitations of vision-based methods~\cite{wei2021real}. Positioned on the upper arm and wrist, in line with best practices~\cite{hoglund2021importance}, the IMUs collect acceleration and orientation data during exercise sets. Repetitions are identified by counting peaks in accelerometer and gyroscope data along the dominant axes for each exercise. This technique, validated in lab tests, shows sufficient reliability and accuracy for estimating repetition counts.

    \subsubsection{Perceived exertion (PE) scoring}
    Our co-design sessions revealed the importance of conversational dialogue during patient-physiotherapist interaction. To accommodate this, our robot asks variations of `How did you find that exercise?', to which the patient provides a verbal response. Our goal is to extract a numerical score in line with the Borg Rating of Perceived Exertion CR10 Scale~\cite{williams2017borg}, a widely validated tool used to measure subjective effort during exercise, where 1=very slight, 3=moderate, 10=maximal. To achieve this, a physiotherapist with eight years of experience labeled 33\% of patient responses. We then used OpenAI’s GPT-4 Large Language Model (LLM) to process patient responses, employing Few-Shot Prompting~\cite{brown2020language} with a Borg Scale explanation and 12 expert-labeled examples. The LLM achieved 94\% agreement with the expert within a ±1 range (48\% exact), tending to rate exertion slightly higher (expert average: 3.06, LLM average: 3.64). This LLM pipeline enables the robot to estimate PE from natural language input (e.g., “That felt fine.”= 3, “That was a little difficult.”= 4, “Still painful and I could do with a drink.”= 7).
    
\section{EXPERT DATA COLLECTION}
    
To develop a behavior policy for our robot instructor, we required a diverse dataset of patient-representative behaviors, capturing variations in responses to exercise exertion and fluctuations in day-to-day performance. However, direct robot-patient interactions at this stage were impractical and posed potential risks without a well-validated control policy. To address this, we leveraged the expertise of healthcare professionals with extensive experience in upper limb rehabilitation and patient behavior. By engaging clinicians as proxies for real patients, we adopted an approach inspired by ‘patient simulation’—a well-established method in medical training where individuals roleplay as patients to facilitate learning and assessment~\cite{javaherian2020role}. Similar techniques, such as \textit{bodystorming}, have been used in HRI research to inform system design where it is otherwise challenging to access real end users~\cite{porfirio2019bodystorming}.

    \subsection{Participants}

    We recruited 33 participants, including 14 physiotherapists, 17 occupational therapists, one assistant practitioner, and one physiotherapy student (with patient practice experience). On average, the participants had 9.9 years of experience in rehabilitating patients with upper limb injuries and came from various public healthcare organizations and geographical locations across Scotland, UK. We held 23 data collection sessions at a local community rehabilitation center gym, with the remainder taking place in academic laboratories across two universities.

    \subsection{Protocol}
    
    
    Each participant completed two exercise sessions guided by our robot prototype, each following the structure outlined in Figure \ref{fig:session-structure}. They were assigned one of 18 scenarios based on a 3x6 matrix of exertion levels (underexert, overexert, optimal exertion) and patient abilities based on variations in ROM and strength levels. Abilities 1-3 would perform \{assistive, active\} phases while abilities 4-6 would perform \{active, resistive\}. Each scenario also included a brief description of the patient’s state. For example, Ability 4 -- Overexertion: \textit{The patient has a high range of motion, performs active exercises well, and is beginning resistive exercises. The robot overexerts the patient by assigning too many repetitions.} The ability descriptions were kept intentionally broad to reflect the diverse patient populations each expert worked with, such as those recovering from fractures or strokes. All participants completed an optimal exertion session, plus either an over or under exerted scenario. 
    
    The robot's actions (e.g., \textit{instruct-assistive-flexion-set-2-reps-4}) were controlled by the researcher via a web interface (Wizard of Oz) while data was recorded through the IMUs as they performed exercises and through a microphone during their verbal responses. Participants were instructed to respond as they would expect patients to during similar points in a session. They were told they could end the session at any time, for instance, if they believed a patient in that particular scenario may be unable to continue due to fatigue. We set an arbitrary number of repetitions per set—six for optimal and overexertion conditions, and four for underexertion.
    
    \subsection{Dataset} \label{sec:dataset}

    \begin{table}
        \centering
        \caption{Exertion tolerance cluster data with assigned groupings}
        \begin{tabular}{ccccccc}
            \hline
            & \multicolumn{2}{c}{\textbf{Under}} & \multicolumn{2}{c}{\textbf{Over}}\\ 
            \textbf{Group} & \textbf{Avg Reps} & \textbf{Avg PE} & \textbf{Avg Reps} & \textbf{Avg PE}\\ 
            \hline
            High Performer & 106.3\% & 1.3 & 86.9\% & 4.4\\
            Average Performer & 115.9\% & 1.8 & 52.9\% & 5.5\\
            Low Performer & 102.4\% & 2.4 & 17.8\% & 6.0\\
            \hline
        \end{tabular}
        \label{tab:exertion-tolerance-table}
    \end{table}

    
    All but one participant completed two robot-guided exercise sessions, with one session excluded due to a recording error, resulting in a total of 63 sessions spanning 16 hours. Session durations ranged from 4 to 21 minutes, with an average of 14 minutes. There were 10, 12, 9, 12, 10, 10 participants for abilities 1-6 respectively, with the following occurrences: Optimal: 32, Underexertion: 16, and Overexertion: 15. 1020 exercise sets were complete (5187 repetitions) and 345 user utterances collected as feedback to the robot's questioning on perceived exertion. 

\begin{figure}[t]
        \centering
        \includegraphics[width=\linewidth]{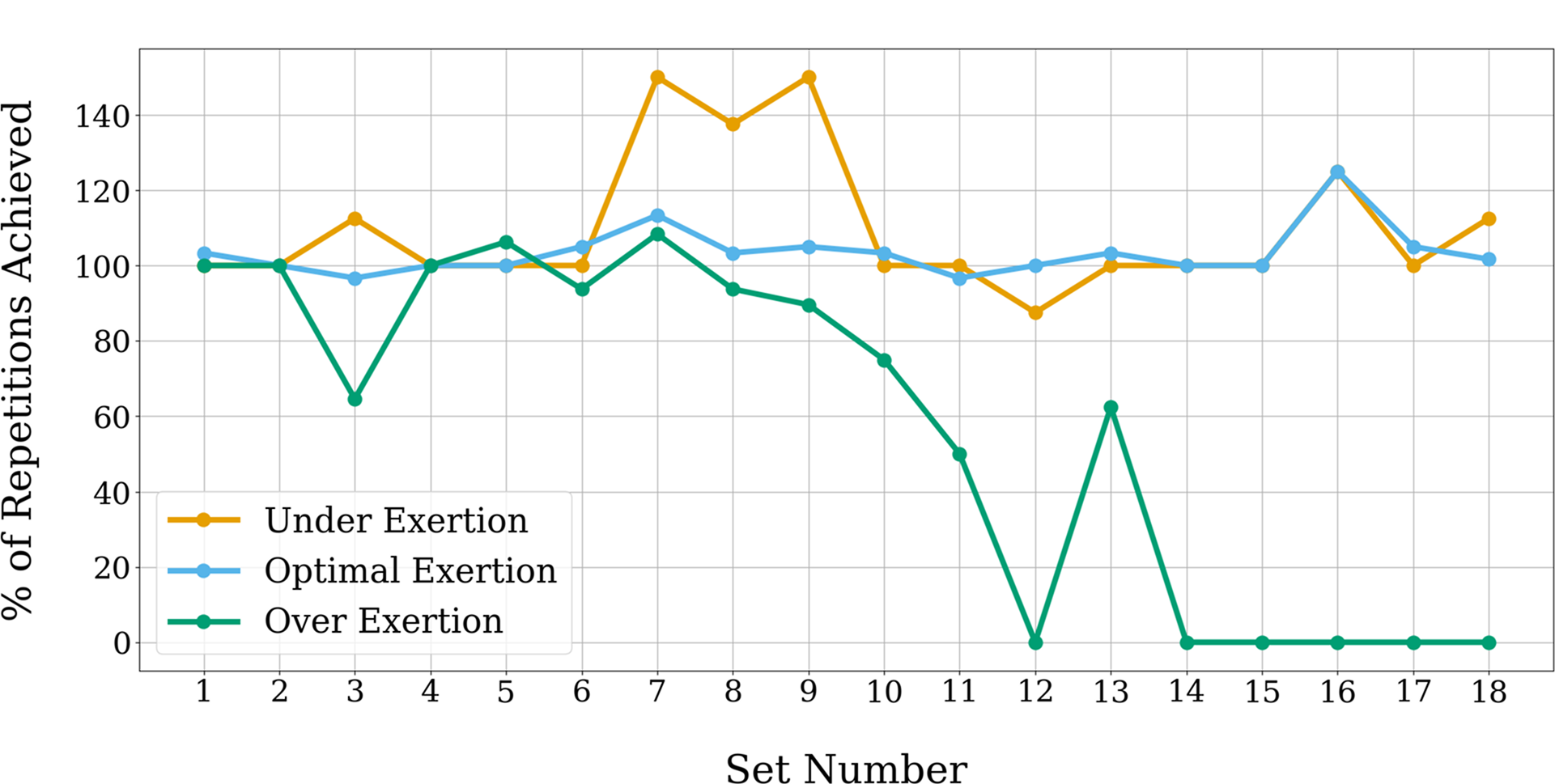}
        \caption{The average number of repetitions achieved (as a \% of number instructed) across all ability level 3 sessions. Each session consists of 18 exercise sets, and the graph illustrates the temporal progression throughout the session.}
        \label{fig:rep-diff-graph}
    \end{figure}

    In line with our robot’s decision-making objectives, we performed K-means clustering (k=3) across average achieved repetitions (as a percentage of instructed) and PE scores for over and under exertion sessions, to uncover variations in how participants responded to different exertion intensities. Some tolerated higher repetitions well, while others showed signs of fatigue or struggled. Table \ref{tab:exertion-tolerance-table} summarizes these the clustered types. Figure \ref{fig:rep-diff-graph} shows an example of the differences across exertion levels for a single ability level. Through analysis of session data and participant feedback, we also identified six common `performance patterns': \textit{struggling day} (consistently below goal), \textit{drastic decline} (sudden drop in performance), \textit{slow decline} (gradual fatigue), \textit{linear} (steady performance), \textit{good day} (above average), and \textit{linear increase} (improving over time).


\section{POLICY LEARNING IN SIMULATION}

    \begin{figure}[t]
        \centering
        \includegraphics[width=\linewidth]{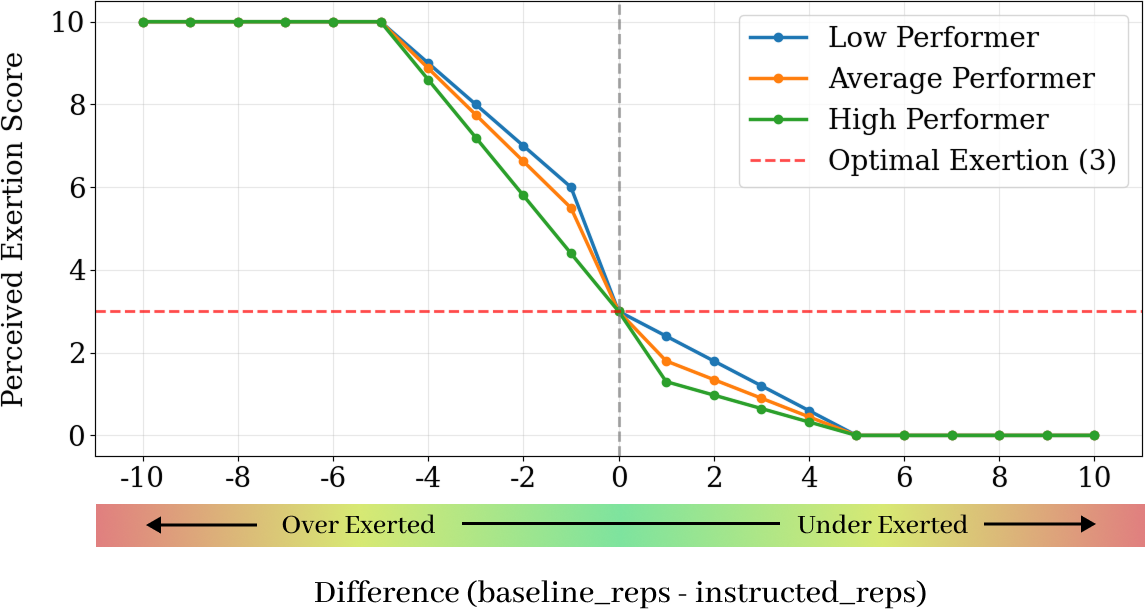}
        \caption{Visualization of the simulation function for perceived exertion score as derived from the repetition difference and the patient's exertion tolerance.}
        \label{fig:pe-graph}
    \end{figure}

    \begin{figure*}[t]
        \centering
        \includegraphics[width=\textwidth]{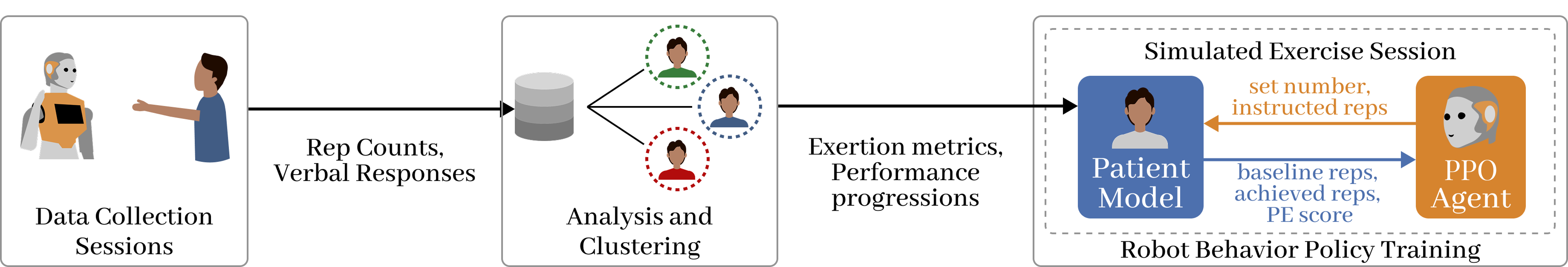}
        \caption{We collect patient-representative behavior from robot-led exercise sessions, process and cluster the data into distinct characteristics, and use these to inform a patient behavior model that interacts with an RL agent for policy learning in simulated sessions.}
        \label{fig:sim-architecture}
    \end{figure*}

We developed a software simulation environment for robot-patient interaction during our exercise sessions. The environment consists of a patient behavior model, informed by our dataset, that manages simulated patients and interacts with a reinforcement learning agent to instruct repetitions. This feedback loop enables policy learning (Figure \ref{fig:sim-architecture}).

    \subsection{Patient Behavior Model}

    Each episode initializes the patient behavior model with a profile containing static and dynamic attributes that govern patient responses to robot instructions. In real-world use, we could imagine known patient attributes of this profile be set by the patient's physiotherapist. For our use, we seed the profile based on identified trends from our collected dataset:
    


    \begin{itemize}
        \item Exercise plan: Goal repetition counts for each of the 18 sets in the session (e.g., 10*18).
        \item Exertion tolerance: Low Performer, Average Performer, or High Performer.
        \item Daily performance: One of six performance pattern variations introduced in Section \ref{sec:dataset}.
        \item Performance tracker: The model also stores patient performance over the live session.
    \end{itemize}
    
    At each set, the patient model receives \textit{set\_number} and \textit{instructed\_reps}, and generates \textit{baseline\_reps}, \textit{achieved\_reps}, and \textit{PE\_score}. \textit{baseline\_reps} is the number of repetitions our model expects the patient to achieve in line with their given daily performance. It is determined by adjusting the goal repetition count for the current set by a function that simulates daily performance with an added noise factor (examples in Section \ref{sec:results}). Equation \ref{eq:achieved_reps} shows how the achieved repetitions are calculated:

    \begin{equation}
    \resizebox{\columnwidth}{!}{$
    \textit{achieved\_reps} =
    \begin{cases}
    \textit{baseline\_reps} & \text{if } \textit{instructed\_reps} \geq \textit{baseline\_reps}, \\
    \textit{instructed\_reps} & \text{if } \textit{instructed\_reps} < \textit{baseline\_reps}, \\
    1 & \text{if } \textit{baseline\_reps} < 1
    \end{cases}
    $}
    \label{eq:achieved_reps}
    \end{equation}

    \textit{PE\_score} is calculated by mapping the difference between \textit{baseline\_reps} and \textit{instructed\_reps} to the Borg CR10 scale. For optimal exertion (\textit{baseline\_reps} = \textit{instructed\_reps}), the score is fixed at 3.0. For underexertion (\textit{baseline\_reps} $>$ \textit{instructed\_reps}) or overexertion (\textit{baseline\_reps} $<$ \textit{instructed\_reps}), the score is based on the magnitude of the difference, adjusted by the patient's exertion tolerance profile from our clustering analysis (Table~\ref{tab:exertion-tolerance-table}). The score is derived by averaging the profile’s exertion range for a 1-rep difference, then scaling linearly to extremes (0 for underexertion, 10 for overexertion) as the difference approaches 5 reps. Differences over 5 reps return the scale's extremes. This relationship is visualized in Figure~\ref{fig:pe-graph}.
    
    \subsection{Reinforcement Learning Agent}


    We formulate our robot's decision-making as a RL problem. RL is well-suited for generating complex control policies through trial-and-error interactions. This involves our robot instructor agent engaging with our patient model in simulated sessions, where it learns to map session states to optimal instruction actions to maximize cumulative session reward~\cite{sutton2018reinforcement}. The resulting policy aims to dynamically adapt its guidance, accounting for patient ability and session variability.

    \subsubsection{State space} 
    
    \(S\) is defined as:

    \begin{itemize}
        \item \( s_{\text{set}} \in \{1, \dots, 18\} \) is the current set.
        \item \( s_{\text{exertion tolerance}} \in \{\text{low}, \text{average}, \text{high}\} \) reflects the category of patient exertion tolerance (taken from the patient's profile).
        \item \( s_{\text{avg reps}} \in \{30-39.9\%, \dots, 140-149.9\%\} \) denotes the running average repetition count category with a window of the past two sets in the session, as a percentage of the patient's goal reps. This tells the agent how the patient's performance is changing over the session.
    \end{itemize}
    
    \subsubsection{Action space}
    
    \( A = \{-70\%, \dots, +70\%\} \) consists of percentage-based adjustments to the goal repetition count for the current set. This formulation allows for adaptive scaling across different exercise plans, ensuring that adjustments remain proportional regardless of exercise plan.
    
    \subsubsection{Reward function}


    \begin{equation}
        R_{\text{total}} = w_1 \underbrace{\left(\frac{\text{reps achieved}}{\text{baseline reps}}\right)^2}_{R_{\text{reps}}} 
        + w_2 \underbrace{\left(1\!-\!\frac{(\text{PE score} - 3)^2}{\alpha} \right)}_{R_{\text{feedback}}}
        \label{eq:reward}
    \end{equation}
    
    \( R_{\text{total}} \), shown in Equation \ref{eq:reward}, is an incentive to guide the agent's learning objectives. In our composite reward function, \( R_{\text{reps}} \) encourages the agent to maximize the number of repetitions achieved by the patient relative to their capability in that set. The quadratic form makes higher repetition counts disproportionately more rewarding. However, without considering the patient's subjective physical state, the agent would be driven to push the patient at the cost of potentially injuring them. To address this, \( R_{\text{feedback}} \) regulates the PE score, ensuring that it remains within a moderate range. \( \alpha \) controls the steepness of the penalty for deviations from the ideal PE of 3. In practice, \(\alpha = 2 \text{ for } \text{PE} < 2; \quad 3 \text{ for } 3 \leq \text{PE} \leq 8; \quad 2 \text{ for } \text{PE} > 8.\) The weights \( w_1 \) and \( w_2 \) (set for us to 0.8 and 0.2 respectively) can be tuned based on experimental results to prioritize the different signals. \( R_{\text{total}} \) is normalized between -1 and 1.
    

    \subsubsection{Proximal Policy Optimization (PPO)}
    
    Given the moderate complexity of the state-action space, we opted for a PPO agent. PPO is a policy gradient method that optimizes a surrogate objective function while ensuring that updates to the policy do not deviate too much from the previous policy. This approach balances exploration and exploitation, allowing the agent to learn stable and efficient policies in high-dimensional environments.

\section{RESULTS} \label{sec:results}

We trained our PPO agent over 100,000 timesteps (5,555 episodes). The agent achieved an average episode reward of 13.8 out of a possible 18, indicating good performance on our simulated sessions with patients. To gain better insight into our learned policy's behavior, we analyzed average performance over 500 exercise session episodes for a variety of patient profile combinations. To simplify visualization, we use a exercise plan of 10 repetitions for each of the 18 sets.

\begin{figure}[t]
    \centering
    \includegraphics[width=\linewidth]{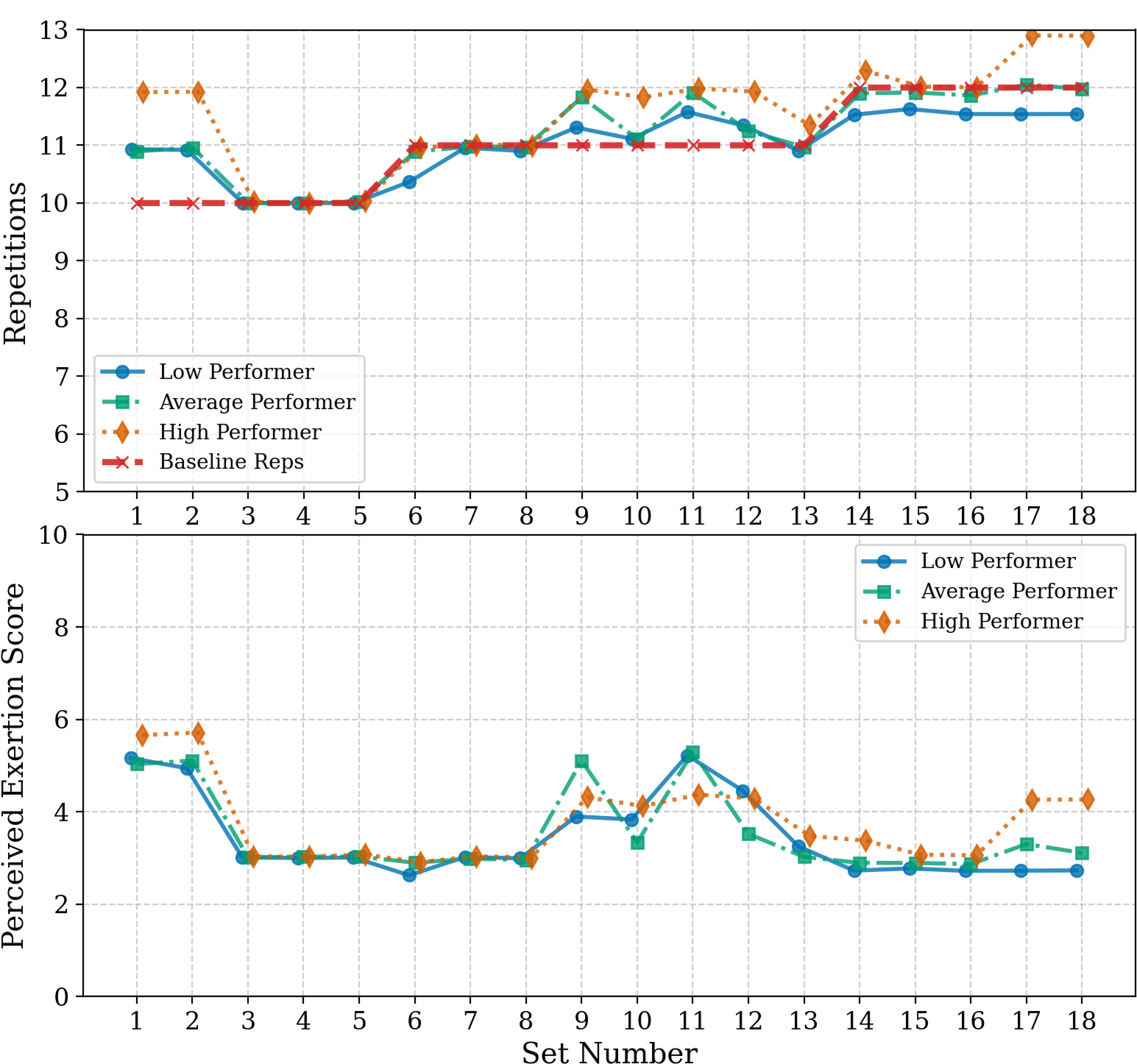}
    \caption{Average performance of our trained policy on \textit{instructed reps} (TOP) and \textit{PE score} (BOTTOM) in sessions patients perform increasingly well.}
    \label{fig:result_linear_all}
\end{figure}

Figure \ref{fig:result_linear_all} illustrates the policy’s performance under the \textit{linear increase} pattern, where patients improve across three exertion tolerance groups throughout the session. \textit{Baseline reps} represent the patient's expected performance as generated from our patient simulation. The policy starts more optimistically for high performers but quickly levels to baseline by set 3 across all groups. It anticipates good performance, gradually increasing the instructed repetition counts, with a noticeable difference between groups in the middle phase—high performers are consistently instructed one more repetition than the others. By the end, the policy gives more relief to low performers while continuing to challenge high performers. The average difference of repetitions counts between baseline and instructed is as follows: Low Performers: 0.01, Average Performers: 0.19, High Performers: 0.56. The policy quickly adjusts the PE score after set 2 and effectively controls it as it peaks again through the middle phase. The average PE for each exertion tolerance in the sessions are: Low: 3.45, Average: 3.52, High: 3.78.

\begin{figure}[t]
    \centering
    \includegraphics[width=\linewidth]{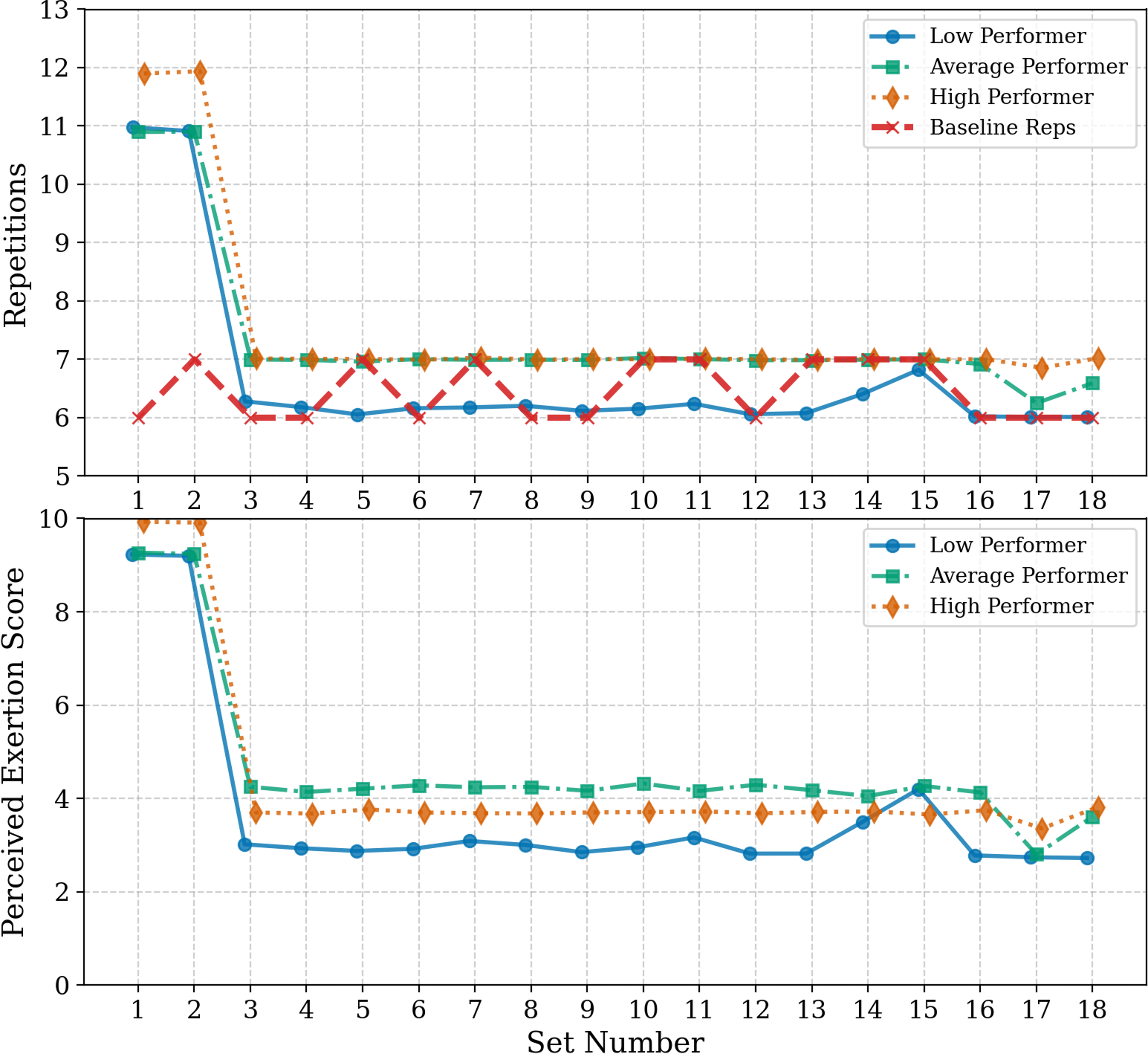}
    \caption{Average performance of our trained policy on \textit{instructed reps} (TOP) and \textit{PE score} (BOTTOM) in sessions patients perform poorly.}
    \label{fig:result_struggling_all}
\end{figure}

Figure \ref{fig:result_struggling_all} illustrates the policy’s performance under the \textit{struggling day} pattern. Initially, the policy applies a similar strategy as observed in the previous example, lacking prior session data. By set 3, it adapts, reducing instructed repetitions to match performance. Low performers experience the sharpest drop at set 3, stabilizing at around 6 reps for the remainder of the session. Average performers and high performers maintain instruction at 7 reps, with average performers receiving minor relief in the penultimate set. The average instructed repetition difference from baseline is 0.27, 0.92, and 1.10 for the three patient exertion tolerances. The average PE score starts high, but is quickly stabilizes (averages: 3.71, 4.66, 4.38).

\begin{figure}[t]
    \centering
    \includegraphics[width=\linewidth]{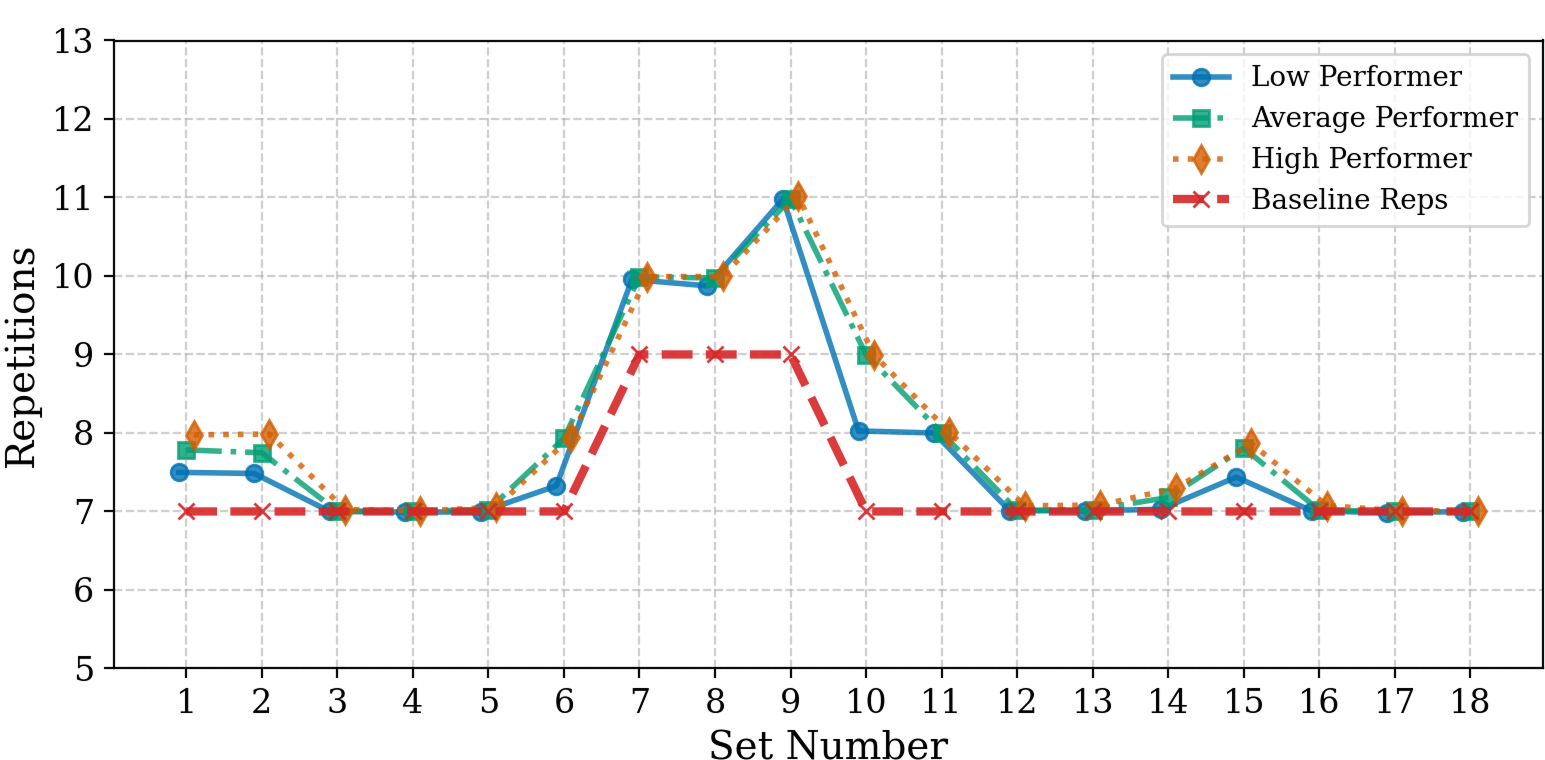}
    \caption{Average performance of our trained policy on \textit{instructed reps} on a more complex exercise plan, with goal repetitions of 7 for all sets besides sets 7, 8, 9, where the goal is 9 repetitions.}
    \label{fig:result_complex}
\end{figure}

Figure \ref{fig:result_complex} demonstrates how our policy performs on a more tailored exercise plan with the \textit{linear} pattern, where patients perform close to their goal repetitions for the duration of the session. In these patients' exercise plans, all sets are 7 repetitions except the first block of rotation exercises (sets 7, 8, 9), which are set to 9 . Notably, the policy maintains appropriate adjustments even when faced with fluctuating repetition goals. The average difference in repetitions from instructed to baseline is 0.43, 0.58, and 0.63 for the three patient exertion tolerances. The PE score also remains controlled (average: 4.17, 4.31, 3.90).

We observed similar performance across the remaining three patterns (\textit{drastic decline}, \textit{slow decline}, and \textit{good day}). Notably, in the \textit{good day} pattern, the policy was slow to increase repetitions for low performers who started strong, with an average repetition difference of -0.76.


\section{DISCUSSION}

The objectives of a our robot instructor's behavior policy were threefold: personalization to individual exertion tolerance, adaptation to varying session performance, and applicability across different exercise plans.
Our results demonstrate that the generated policy has learned distinct approaches in session instruction across three patient exertion tolerance groups learned from our dataset, pushing high performers with greater repetition instructions while instructing low performers more conservatively. Results also show that the policy adapts to different performance patterns, recognizing when patients are performing well and adjusting repetitions accordingly, or quickly reducing them when patients struggle. Lastly, we observe our policy working on a more customized exercise plan with varying repetitions counts per exercise block.

Our work presents two key novelties. First, our RL-based framework enables adaptive instruction beyond existing social robot exercise instructors, dynamically adjusting to patient exercise performance sensed through IMUs and naturalistic verbal feedback through an LLM-based approach. By incorporating nine exercise variations and adapting to different exercise plans, our system also promotes usability across a wide range of patient abilities. Second, unlike prior RL-based HRI studies that rely on generalized user behavior models~\cite{stolarz2024learning, tsiakas2018task}, we integrate expert knowledge of patient rehabilitation, replacing naive user assumptions with informed models—bringing us closer to effective real-world robot behavior.

    \subsection{Limitations and Next Steps}

    Perceived Exertion (PE) is highly subjective and unreliable as a sole indicator of exercise difficulty; future work should incorporate objective metrics such as Range of Motion to better assess fatigue. The current policy lacks awareness of patient performance prior to each session, often leading to overexertion when patients are fatigued—this could be mitigated through an initial assessment phase. Although PE is integrated via naturalistic voice interaction, translating open-ended verbal input into numerical scores remains challenging due to irrelevant responses and microphone reliability. Future iterations will incorporate additional input modalities to improve PE interpretation. Model generalizability to unseen behavior trajectories is likely limited by the current rule-based patient model; future work will explore more dynamic alternatives, potentially leveraging LLMs. The existing repetition counting pipeline from IMU data is not generalizable to a wider range of exercises. Future work could look to use more sophisticated machine learning approaches to strengthen this. Finally, future work will focus on integrating multi-session learning to track patient progress over time and refine personalization, allowing the robot to support long-term therapeutic relationships rather than adapting only within single sessions.

\section{Conclusion}

In this work, we present a social robot exercise instructor for standardized upper limb exercises, building on prior research~\cite{bettosidesigning, bettosi2024systematic}. To address data limitations, we used 33 expert clinicians as proxies to create a diverse dataset of patient-representative behaviors. Based on this, we developed a patient behavior model and trained a RL agent to personalize exercise instructions, adjusting repetition counts according to exertion tolerances and session-to-session performance fluctuations. Our results show promising adaptation in simulation, laying the foundation for autonomous, personalized exercise guidance beyond current implementations. Our next steps are focused on validating the system through real-world patient evaluations and creating further enhancements with multi-session learning and expanded perception capabilities.



\section*{ACKNOWLEDGMENTS}

We would like to thank South Lanarkshire University Health and Social Care Partnership, and the staff at both NHS Lanarkshire and Blantyre LIFE for their valuable input during the data collection phase of this work. We would also like to thank Tom McKeever for his clinical input on this paper.


\bibliographystyle{IEEEtran}
\bibliography{references}

\end{document}